\documentclass{article}

\pdfoutput=1 

    \PassOptionsToPackage{square,sort,comma,numbers}{natbib}

    \usepackage[preprint]{neurips_2019}

\usepackage[utf8]{inputenc} 
\usepackage[T1]{fontenc}    
\usepackage{hyperref}       
\usepackage{url}            
\usepackage{booktabs}       
\usepackage{amsfonts}       
\usepackage{nicefrac}       
\usepackage{microtype}      
\usepackage[mathscr]{euscript}

\makeatletter  
\newif\if@restonecol  
\makeatother

\usepackage[linesnumbered,ruled,vlined]{algorithm2e}
\usepackage{algpseudocode}  
\usepackage{amsmath}  

\usepackage{subfigure}
\usepackage{graphicx}
\usepackage{algpseudocode}
\usepackage{amsmath}
\usepackage{amsthm}

\usepackage{graphicx}
\usepackage[alpha]{mdpn}

\title{Learning Efficient and Effective Exploration Policies with Counterfactual Meta Policy}

\author{%
    Ruihan Yang\thanks{equal contribution} \\
  Nankai University\\
   \And
   Qiwei Ye \footnotemark[1]\\
   Microsoft Research Asia \\
   \And
    Tie-Yan Liu \\
   Microsoft Research Asia \\
}

\theoremstyle{definition}
\newtheorem{definition}{Definition}[section]

\begin{document}

\maketitle

\begin{abstract}

A fundamental issue in reinforcement learning algorithms is the balance between exploration of the environment and exploitation of information already obtained by the agent. Especially, exploration has played a critical role for both efficiency and efficacy of the learning process. However, Existing works for exploration involve task-agnostic design, that is performing well in one environment, but be ill-suited to another. To the purpose of learning an effective and efficient exploration policy in an automated manner. We formalized a feasible metric for measuring the utility of exploration based on counterfactual ideology. Based on that, We proposed an end-to-end algorithm to learn exploration policy by meta-learning. We demonstrate that our method achieves good results compared to previous works in the high-dimensional control tasks in MuJoCo simulator. 


\end{abstract}
\section{Introduction}
Deep reinforcement learning (RL) has succeed in multiple areas recently, including robotics~\citep{gu2017deep}, continuous control~\citep{lillicrap2015continuous}, and video games~\citep{tian2017elf,mnih2015human}. Typically, the RL agent goes through a process of balancing between exploration of the environment and exploitation of policies already learned. The agent has to exploit what it has already experienced in order to obtain better rewards, but it also has to explore in order to obtain a better view of the environment that may yield better future rewards. While there have been a lot of works addressing how to perform efficient and effective exploitations, such as TRPO and PPO~\citep{schulman2015trust, schulman2017proximal}, the efforts on understanding and improving explorations is yet insufficient. 

Actually, exploration plays a very critical role in reinforcement learning. On one hand, exploration gathers the information about environments, so that the RL agent could have the evidence and confidence to maximize the long-term reward even if it has to sacrifice some short-term rewards. It is clear that a better exploration would enable the RL agent to make more informed tradeoff between show-term and long-term rewards. On the other hand, a bad exploration strategy could hurt the efficacy and efficiency of exploitation, especially for the off-policy algorithms like DDPG~\citep{lillicrap2016continuous}, DQN~\citep{mnih2013playing}, and SAC~\citep{haarnoja2018soft}. For example, the common exploration strategy for DDPG is to add a Gaussian noise to policies, however, the Gaussian noise may be sub-optimal and may lead to a unfavorable consequence of slowing down the effective exploration. 

The existing attempts on finding better exploration strategies that render more stable and faster policy learning can be classified into two categories. The first kind of works are heuristic, or involve task-agnostic objectives. Examples include intrinsic motivations \citep{schmidhuber1987evolutionary,bellemare2016unifying,singh2010intrinsically,hester2012intrinsically}, uncertainty reduction \citep{houthooft2016vime,lopes2012exploration,tang2017exploration,martin2017count,pathak2017curiosity}, information gain \citep{eysenbach2018diversity,nikolov2018information}, Thompson sampling \citep{chapelle2011empirical,osband2016deep}, network optimization \citep{burda2018exploration,plappert2017parameter,fortunato2017noisy,stadie2015incentivizing,achiam2017surprise}, and expert exploration guided strategies\citep{sequeira2014learning,stadie2015incentivizing}. The second kind of works try to learn the exploration strategies in an automated manner. For example, \cite{gupta2018meta} proposed a fast adaptation algorithm that could leverage prior experience across tasks for exploration strategies based on MAML~\citep{finn2017model}. \cite{sharaf2019meta} introduced an algorithm for learn heuristic exploration strategy that could generalize to future contextual bandit problems by imitation learning. \cite{xu2018learning} proposed a learnable meta policy for exploration towards efficacy. However, as far as we know, there is no work yet that explicitly characterizes both efficacy and efficiency of exploration strategies, and adopts a fully automatic process to learn an optimal exploration policy. This is exactly the motivation of our work.

To this end, we need to have an explicit measure of the quality of information that the agent could extract from the exploration, as well the cost that the agent has to pay. What we have done in this work is to use the counterfactual idea to model the relation between exploration and exploitation. In particular, we introduce a view that decouples exploration strategy as a standalone meta-MDP from the original MDP, based on which, we formalize both {\em gain} and {\em cost} of exploration to characterize its nature. Then, we design an end-to-end algorithm to learn an exploration policy with the optimal tradeoff between gain and cost in a fully automatic manner by means of meta-learning. We apply our proposed method to off-policy algorithms and demonstrate its efficacy and efficiency on Mujoco enviroments~\citep{todorov2012mujoco}. Our method can boost the performance of the vanilla off-policy DRL algorithms by a large margin, mainly because it could provide a more suitable exploration strategy during training than previous works.

The paper is organized as follows. Section \ref{sec::preliminaries} discusses the related works and background. Section \ref{sec::l2e} describes our proposed algorithm for learning exploration policies based on the counterfatual idea. In Section \ref{sec::experiments}, we demonstrate that the proposed algorithm can incorporate modern off-policy algorithms \citep{lillicrap2015continuous} and achieve promising experimental results . We conclude with ablation studies, discussions and commentary on future research directions.

\section{Related Works} \label{sec::preliminaries}
\textbf{Towards more sample efficient exploration strategies.} Traditional, to address the sample efficiency issue, there are few successful exploration strategies works well, such as $\epsilon$-greedy~\citep{sutton1996generalization}, $\epsilon$-decreasing \citep{sutton1998introduction}, Gaussian exploration~\citep{lillicrap2015continuous}, upper confidence bounds \citep{li2010contextual}, exponential gradient $\epsilon$-greedy~\citep{li2010exploitation}. Beyond that, a large family of RL algorithms has been proposed recently based on intrinsic motivations \citep{schmidhuber1987evolutionary,bellemare2016unifying,singh2010intrinsically,hester2012intrinsically}, uncertainty reduction \citep{houthooft2016vime,lopes2012exploration,tang2017exploration,martin2017count,pathak2017curiosity}, information gain \citep{eysenbach2018diversity,nikolov2018information}, Thompson sampling \citep{chapelle2011empirical,osband2016deep}, network optimization \citep{burda2018exploration,plappert2017parameter,fortunato2017noisy,stadie2015incentivizing,achiam2017surprise}, and expert exploration guided strategies\citep{sequeira2014learning,stadie2015incentivizing}. However, most of these algorithms have limited applicability, considering their task agnostic, when applying to rich environments that reinforcement learning tasks facing on. Devising more effective and generalized exploration strategy is, therefore, a critical challenge in reinforcement learning.

For fast adaption to learning different tasks of learning strategies, the idea of meta learning, or learning to learn, has been studied for decades~\citep{schmidhuber1987evolutionary,hochreiter2001learning,thrun2012learning}. 
To our knowledgement, the most related work of ours are \cite{xu2018meta,gupta2018meta,xu2018learning,sharaf2019meta}. \cite{xu2018meta} proposed a  meta-learner, which could learn to turn hyper-parameters of return function, to learn directly by meta learning to maximize the accumulate reward during a single lifetime of interactions with task environment. \cite{gupta2018meta} introduced a algorithms that could learn task specific exploration strategies using prior experience based on MAML, comparing to learning exploration strategy for various of environments, we are more interesting in learn exploration strategy that improve effiecncy during learning. Our approach is most closely related to the work of \cite{xu2018learning}, which has proposed a framework that improves sample-efficiency of DDPG on various of different tasks by offering a meta-reward for exploration policy. However, the exploration not only play the role for better efficacy but also take efficiency into account. Our approach is trying to find the most efficient exploration policy to generating samples that could potentially maximizing the expect reward of agent during off-policy training. 

\textbf{Multi-agent Reinforcement Learning}.From the perspective of multi-agent reinforcement learning, we would describe relationship between exploration agent, decoupled from original agent, and exploitation agent as {\em coopetitive (cooperative \& competitive)}, that is to say the exploration agent not only need help exploitation agent obtaining rewards ({\em Gain}) as much as possible, but also need taking less {\em opportunity cost}. One of critical challenge in multi-agents settings is to do credit assignment among multiple agents~\citep{chang2004all}, especially along with the larger number of agents. Based on the idea of {\em different reward}~\citep{wolpert2002optimal}, \cite{foerster2018counterfactual} has proposed COMA, which core idea is that using counterfactual baseline that will  marginalizes out a single agent’s action, shown as E.q.\eqref{eq:coma}.
\begin{equation} \label{eq:coma}
    A^a(s,u) = Q(s,u) - \sum_{u^a}\pi(u^a|\tau^a)Q(s,(u^{-a},u^a))
\end{equation}
Inspired by the same idea, we formalized {\em Counterfatual Value} to characterize the benefit/cost from exploration and exploitation. Based on that, We have proposed an way to  balance exploration and exploitation implicitly. For more details, please refer to Secion \ref{sec::counterfactual}

\section{Methods} \label{sec::l2e}
Exploration \& exploitation dilemma has be the fundamental issues in Reinforcement Learning. While previous works shows that learning could benefit both efficiency and efficacy from good exploration strategy, the prior methods generally has two major shortcomings: (1) The heuristic based exploration strategy couldn't adapt to the rich environment/problem settings. (2) Meta-learning based exploration strategy doesn't characterize the relationship between exploration and exploitation. In this section, introduce a novel method for learning efficient and effective exploration strategy based on the counterfactual idea by meta-learning.

To answer the questions posed by \cite{thrun1992role}, this proposed method has introduce the view that decouples exploration strategy as a standalone meta-MDP from the original one, then we training the exploration policy towards the direction that could using less {\em opportunity cost} to generate trajectory has large {\em information gain}. Where {\em opportunity cost} and {\em information gain} are defined under meta-MDP. 

\subsection{Meta Markov Decision Process}
MDP is given by $\simpleMDP$, with $\sset$ being the set of states, $\aset$ is the set of actions, $\Tdef$ is called the {\em transition function}. For all $(s,a,s',t) \in \sset \times \aset \times \sset \times \NZ$, let $\Tfun(s,a,s')\coloneqq\Pr(\st{t+1}=s' \mid \st{t}=s, \at{t}=a)$. $\Rfun$ is the reward function. In deep reinforcement learning, the value function and policy are approximated by a neural network
with parameters $\theta$, denoted by $v_\theta(s)$ and $\pi_\theta(a|s)$ respectively. Where the performance of policy function could be measured by:
\begin{equation} 
    J(\pi_{\theta}) = E_{\tau}[G(\tau) | \tau(s_0) \sim \rho, \tau \sim \pi_\theta]
\end{equation}
where $\rho$ is the initial state distribution, $G$ is the return function: $G(s_t) = \sum_{i=0}^{\infty}\gamma^{i} R_{t+i}$.
Then we can update policy function through updated function. e.g. policy gradient theorem~\citep{sutton2000policy}:
\begin{equation} \label{eq:pgt}
    \frac{\partial J(\pi_\theta)}{\partial \theta} = E_{s_t \sim P(\cdot | s_{t-1}, a_{t-1}), a_t \sim \pi_{\theta}(\cdot | s_t)}[\nabla_{\theta} \text{log}\pi_{\theta}(s_t, a_t) Q^{\pi_{\theta}}(s_t, a_t)]
\end{equation}

or Q-learning~\citep{watkins1992q}
\begin{equation}\label{eq::q-learning}
Q(s', a) \gets (1 - \alpha) \cdot Q(s, a) + \alpha \cdot (r + \gamma \cdot \max_{a'} Q(s', a'))
\end{equation}

To view exploration agent as a meta-agent that would generating {\em valuable} samples for policy iteration, we can define meta MDP as follow (we using Euler Script font for distinguish): 

\theoremstyle{definition}
\begin{definition}{Meta-MDP:}\label{def::meta-mdp}
For a {\em exploration agent}, the Markov decision process(MDP) is given by $\metaMDP$, with $\metasset$ being the set of (exploitation) policy $\pi$, $\metaaset$ is the set of $\tau$ which was generated under the exploration policy $\pi_e$, $\metaTdef$ is the {\em policy updater transition function}, that is, the transition function that describe  $\pi$ is updated under $\tau$. %
$\metaRfun$ is the reward function for exploration policy. 
\end{definition}

Where $\pi_e$ is exploration policy (parameterized by $\phi$), similar to the work of \cite{xu2018learning}, we can define the meta-reward $\mathscr{R}$ as the performance improvement of different version of policy $\mathscr{R} = {R}_{\pi'} - {R}_{\pi}$, where $\pi'$ is trained from $\pi$ by policy updater using $\tau$

\begin{definition}{Meta-$\mathscr{Q}$:}\label{def::meta-q}
Formally, we can have meta Q-function $\mathscr{Q}$ to describe the expected meta-return that meta-agent could obtained in meta episode, by taking specific action $\hat{a} \in \metaaset$ in given state $\hat{s} \in \metasset$
\begin{equation}
    \mathscr{Q}(\hat{s}, \hat{a}) = \mathscr{Q}(\pi, \tau^t) = \sum_{i=0}^{\infty}{\gamma^{i} \mathscr{R}_{t+i}} 
\end{equation}
\end{definition}

Particularly, the meta Q-function would be described as the expectation of future performance gain of exploitation policy (meta-return for brevity) under the action of using a certain trajectory $\tau$ for policy iteration.

\subsection{Counterfactual Value For Exploration} \label{sec::counterfactual}
Counterfactual idea is a concept firstly introduced in psychology describing the human tendency to think about what not actually happened and its influence on the reality. In general reinforcement learning, advantage function $A(s,a) = Q(s,a) - V(s)$, which is always used in actor-critic framework, is one of the instance of counterfactual ideology.
$V(S)$ can be view as the average return in state $S$. That is to say the advantage function is to describe the estimate return of agent taking action $a$ instead of "average" action.

Intuitively, the metric of the (potential) influence of exploration policy could be decoupled into two aspect: \textbf{Gain:} the expected meta-reward and expected reward if using exploration policy for trajectory $\tau_{\pi_e}$.\textbf{Cost:} The expected meta-reward and expected reward if using action from exploration policy .

Given a formalized meta MDP (See.\ref{def::meta-mdp}), we employed the counterfactual ideology to measure the performance of a exploration policy, We have:
\begin{definition}{\textit{Counterfactual Value:}}\label{def::cv}
Counterfactual Value(CV) is the metric to measure the utility of exploration policy $\pi_e$ for exploitation policy $\pi$ under certain trajectory $\tau$.
\begin{equation}
\begin{split}
    Gain(\pi, \pi_e, \tau ) \vcentcolon&= \mathscr{Q}( \pi, \Bar{\tau}_{\pi_{e}} ) + \sum_{s \sim \tau } Q( s, \pi_e(a))    \\ 
    Cost(\pi, \pi_e, \tau ) \vcentcolon&= \mathscr{Q}( \pi, \Bar{\tau}_{\pi} ) + \sum_{s \sim \tau } Q( s, \pi(a)) \\  
    CV(\pi, \pi_e, \tau ) \vcentcolon&= Gain(\pi, \pi_e, \tau ) - Cost(\pi, \pi_e, \tau ) 
\end{split}
\end{equation}
where $ \Bar{\tau}_{\pi} = \{ (s , a) | s \sim \tau, a=\pi(s) \} $.
\end{definition}

We would like the exploration learning towards the direction that maximize performance gain of exploitation policy efficiently. By the help of definition \ref{def::cv}, we could naturally derive the objective for the exploration policy as:
\begin{equation}\label{eq::metaAdvantage}
\begin{split}
    \hat{A}(\pi, \pi_e, \tau) \vcentcolon&= CV(\pi, \pi_e, \tau) \\
    &= \mathscr{Q} ( \pi, \Bar{\tau}_{\pi_{e}} ) - \mathscr{Q} ( \pi,\Bar{ \tau}_{\pi} ) + \beta \sum_{s \sim \tau } ({Q(s, \pi_e(a)) - Q(s, \pi(a)))}
\end{split}
\end{equation}
Thus, adapting from E.q \eqref{eq:pgt}, we have:
\begin{equation}\label{eq::metapgt}
\begin{split}
\partial J(\phi) &= E [\nabla_{\phi} \text{log}\dot{\Pr}(\hat{a}_t|\hat{s}_t)\hat{A}(\pi, \pi_e, \tau)] \\
&= E [\nabla_{\phi} \text{log}\dot{\Pr(s_1)}\prod_{t=1}^{T}\pi_e(a_t|s_t)\Pr(s_{t+1}|s_t,a_t)\hat{A}(\pi, \pi_e, \tau)] \\
& = E [\sum_{t=1}^{T} \nabla_{\phi} \text{log}\pi_e(a_t|s_t)\hat{A}(\pi, \pi_e, \tau)]  
\end{split}
\end{equation}

where the $\dot{\Pr}(\hat{a}_t|\hat{s}_t)$ is the probability of trajectories $\tau$ ($\hat{a}_t$) was generated under exploration policy $\pi_{e}$ ($\hat{s}_t$). For the efficiency consideration, we introduce one hyper-parameter $\beta$ for balance between exploration and exploitation.

\subsection{Counterfactual Meta Policy}\label{sec::l2ecm}
\begin{figure*}[t]\label{fig::relationship}
    \centering
    \includegraphics[width=0.5\textwidth]{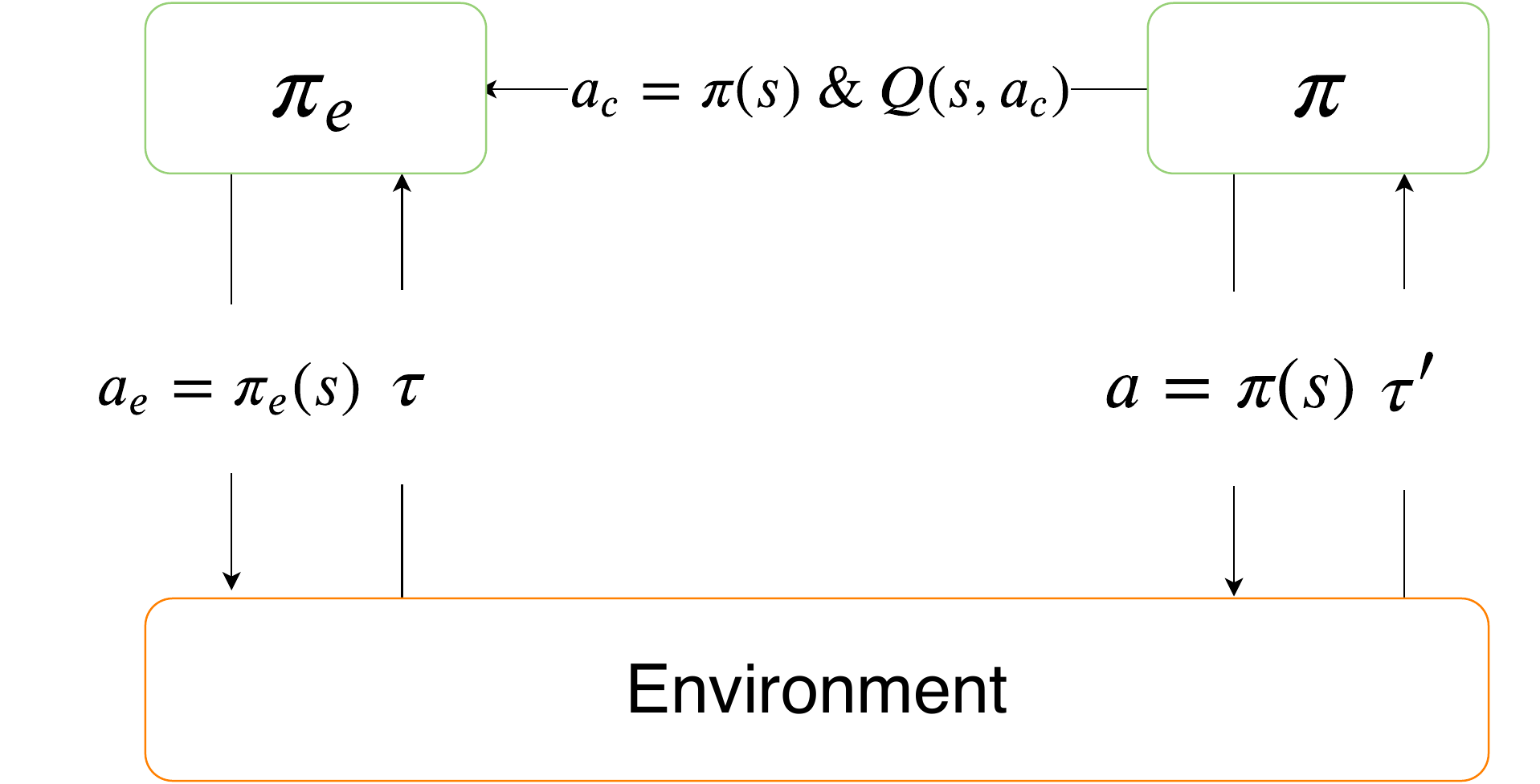}
    \caption{ Overview of Exploration and Exploitaion policies: Exploration policy $\pi_e$ interact with environment and collect exploration samples $\tau$ to improve the performance of exploitation policy. Exploitation policy $\pi$ interact with environment and collect exploitation samples $\tau'$ to evaluate the performance of the policy $\pi$ to compute meta reward $\mathscr{R} = {R}_{\pi'} - {R}_{\pi}$. Exploration policy $\pi_e$ consult the exploitation policy $\pi$ for the counterfactual actions $a_c$ to compute the training objective from E.q \eqref{eq::metaAdvantage}  }
\end{figure*}

The overview of the relationship between our exploration policy and existing exploitation policy is shown in Figure. \ref{fig::relationship}, to update the meta-policy, we could using vanilla REINFORCE with loss function E.q \eqref{eq::metapgt}. What's more, to serve the efficiency manner, we employ actor-critic framework and replace the $\mathscr{Q}$ value in E.q \eqref{eq::metaAdvantage} to reduce the variance.

The computation complexity is high for the exploration policy's optimizing, because of E.q \eqref{eq::metaAdvantage} rely on the estimation of $ \mathscr{Q}( \pi, \tau_{\pi_{e}} ) $ and $ \mathscr{Q}( \pi, \tau_{\pi} ) $.  

According to meta Q-function's definition (See \ref{def::meta-q}), the meta Q-function is defined over policy and trajectories by certain policy. However it will be extremely inefficient to use parameterized model to predict value of meta Q-function using policy and trajectories as direct input. Thus, we employ a function $ \hat{\mathscr{Q}}_{\pi_{t}}$ (parameterized by neural network $\eta$) to estimate the meta Q value by learning the manipulation from each (state, action) pair in that trajectories, and use the estimate value to represent the true meta Q value under certain policy.

\begin{equation}\label{'eq:MetaQ'}
    \mathscr{Q}(\pi, \tau_{\pi'}) \approx \hat{\mathscr{Q}}_{\pi}(s,a ), (s,a) \sim \tau_{\pi'}
\end{equation}
In this case, function $\hat{\mathscr{Q}} $ is trained at the $t$th epoch could be denoted as $\hat{\mathscr{Q}}_{\pi_{t}}$. Following the E.q \eqref{eq::q-learning}, we have:

\begin{equation}
    \mathscr{Q}\left( \pi_{t}, \tau_{t} \right) \leftarrow \mathscr{Q}\left(\pi_{t}, \tau_{t}\right)+\alpha\left(\frac{ \mathscr{Q}\left(\pi_{t-1}, \tau_{t-1}\right) - \mathscr{R}_{t-1} }{\gamma} -\mathscr{Q}\left(\pi_{t}, \tau_{t} \right)\right)
\end{equation}

Note that we are learning the gap of $\hat{\mathscr{Q}}_{\pi_{t}}$ and $\hat{\mathscr{Q}}_{\pi_{t-1}}$.

Thus, along with E.q.~\eqref{'eq:MetaQ'}, we define the loss function for $\hat{\mathscr{Q}}$:

\begin{equation}\label{eq:MSE}
    \mathscr{J}(\hat{\mathscr{Q}}_{\pi_t} ) = ( \sum_{(s,a) \sim \tau_{t} } \hat{\mathscr{Q}}_{\pi_{t}}(s,a) -  \frac{ \sum_{(s,a) \sim \tau_{t-1} } 
    \hat{\mathscr{Q}}_{\pi_{t-1}}\left(s,a\right) - \mathscr{R}_{t-1} }{\gamma}  )^2
\end{equation}
where $\mathscr{Q}_{{\pi_t-1}}$ and $\tau_{t-1}$ are stored at previous iteration. By now, each term of E.q.~\eqref{eq::metapgt} can be computed efficiently with parameterized model, based on actor-critic framework.





\begin{algorithm}\label{alg::cmp}
  \caption{Learning to Explore with Counterfatual Meta Policy}  
    initialize $\pi$, $Q$, $\pi_e$ and $\hat{\mathscr{Q}}$ with $ \theta $, $ \theta' $, $ \phi $ and$ \phi' $ 
    get target $\hat{\mathscr{Q}}'$ \\
    Draw $\tau'^{0}$ with $\pi$ to estimate the performance $ R_{\pi}$ of $\pi$ \\
    initialize replay buffer B using $\tau'^{0}$ \\
    \For{$i=1;i \le n;$}  
    {  
        Generate exploration $\tau^{i}$ with exploration policy $\pi_e$ \\
        Update the exploitation policy $\pi$ to $\pi'$ using Updater with $\tau^i$ \\
        Evaluate policy $\pi'$, get performance $R_{\pi'}$ of $\pi'$ and evaluation dataset $\tau'_{i}$ \\
        Compute reward $ \mathscr{R}(\tau_{i}) = R_{\pi'} - R_{\pi} $  \\
        \For {$\pi_e$ Update times}
        {
            \For{ $\hat{\mathscr{Q}}$ update times}
            {
                get $\mathscr{Q}$ target using $\hat{\mathscr{Q}}'$ \\
                Update $\hat{\mathscr{Q}}$ with E.q \eqref{eq:MSE} \\
                Soft update $\hat{\mathscr{Q}}'$ \\
            }
            Compute $\hat{A}(\pi, \pi_e, \tau)$ \\
            Update exploration policy $\pi_e$ with \eqref{eq::metapgt}; \\
        }
        Put samples into the replay buffer: $B = B \ \cup \ \tau^i \ \cup \tau'^i$ \\
        \For{ Updater update times }
        {
            Update $\pi$ with Updater and $B$ \\
        }
        Evaluate policy $\pi$, get performance $\hat R_{\pi}$ of $\pi$; \\
    }
    return $con(r_i)$\;  
\end{algorithm}

The full procedure of our method is shown in Algorithm \ref{alg::cmp}. In practice, because of the approximation nature of $\hat{\mathscr{Q}}$, we often taking more times than to update than $\pi_e$. We using the loss of E.q \eqref{eq::metapgt}, by maximizing $\hat{A}(\pi, \pi_e, \tau)$, intuitively, the fist term $\mathscr{Q}( \pi, \Bar{\tau}_{\pi_{e}} ) - \mathscr{Q} ( \pi, \Bar{\tau}_{\pi} )$ is indicate that exploration should be optimized towards the direction that long-term (potential) reward could be obtained. And the second term $\sum_{s \sim \tau } {(Q(s, \pi_e(a)) - Q(s, \pi(a)))}$ indicate the sacrifice, considering $Q^*(s, \pi_e(a) <= Q^*(s, \pi(a))$.

Though the meta exploration policy and meta $\mathscr{Q}$ function are introduced, In the contrast of these model taking computing resources to be learned, the training efficiency is actually being improved, for the reason that comparing to the heuristic exploration strategy, samples collected by the exploration strategy are much more suitable for policy iteration to improve the performance. An additional experiment are conducted in Section \ref{sec::experiments} to demonstrate our methods is more efficient given the same amount of time and computing resources.

\subsection{Exploration \& Exploitation Balance with Counterfactual Value} \label{sec::discussion}

The Counterfactual Value, by design, has offered one way that serves the quantitative metric for exploration well. Because of this cost-effective definition, the tradeoff between Exploration \& Exploitation could learn implicitly. However, the nature of that our method always using exploration policy to generate samples for exploitation also need to be discussed. To enable the tradeoff between exploration and exploitation. $\beta$ has been introduced, the larger of the $\beta$, or precisely,  $ \sum_{s \sim \tau } {(Q(s, \pi_e(a)) - Q(s, \pi(a)))}$   > $\mathscr{Q}( \pi, \Bar{\tau}_{\pi_e} ) - \mathscr{Q}( \pi, \Bar{\tau}_{\pi} ) $, has been chosen, the exploration will tend to be like exploitation policy,  and vice versa. We have demonstrated $\beta$ has the ability to control the balance. A clear future work of our methods is to find a better way to tune $\beta$ to adapt to different tasks.

Here we give a shallow analysis of this learned exploration policy. Even our methods have defined a clear optimizing objective for exploration, the role of exploration still remains unclear.  Our design principle is that exploration should always help to exploit better. However, for the meta-MDP of exploration, the definition of meta-reward, which defined the 'information gain' for exploration, could be varied under different perspectives. Another following work will be to have a further understanding of the relationship between exploration and exploitation.

\section{Experiment}\label{sec::experiments}

In this section, we present the detailed experiment results to discuss the overall performance of our algorithm and the use of each part of our algorithm. 
 
Our meta exploration algorithm is compatible with all off-policy methods and we choose DDPG, one of the most popular off-policy methods for continuous control tasks, serves as the baseline. We conduct the experiments on the suite of Mujoco~\citep{todorov2012mujoco} continuous control tasks, provided by OpenAI Gym~\citep{brockman2016openai}. Specifically, to ensure reproducible comparison,  we tested our algorithm on the set of tasks mostly from \citep{xu2018learning}(denoted as L2Explore for brevity), consist of HalfCheetah-v2, Pendulum-v0, Hopper-v2, Walker2d-v2, InvertedPendulum-v2, and InvertedDoublePendulum-v2. All of our experiments ran on a machine equipped with Intel(R) Xeon(R) CPU E5-2690 v4 @ 2.60GHz and two Nvidia Tesla P40 GPUs for ten times using different random seed.

\subsection{ General Experiment setting}

For implementation, Our deterministic actor policy, q network and the meta Q network share the same network architecture with the OpenAI baseline\footnote[1]{https://github.com/openai/baselines} which is a two-layer MLP contains 64 and 64 hidden units with ReLU non-linear activation function and layer normalization~\citep{lei2016layer}. Our meta exploration policy is made up of an MLP containing two hidden layers(64-64) and a vector representing the log-standard-deviation for actions.

To make a fair comparison with our baseline methods: DDPG and L2Explore, we share most of the hyper-parameter as described in the L2Explore or set in the OpenAI baseline. 
Some of the important hyper-parameters are listed here: exploration step {100,1000}, evaluation step {200, 2000}, exploitation policy update times {50, 500}, exploration policy update times {50,500}, considering there are environments has longer episode and need more rollouts, e.g. HalfCheetahv2, comparing to others.

For the hyper-parameter introduced the algorithm, we set meta Q function update times as $5$ for all environments, $\beta$ are discussed in the next section.

\subsection{ Ablation study }

HalfCheetah is one of the most well-known continuous control environment with 17-dimensional observation space and 6 dimensional. HalfCheetah needs a good exploration strategy, where the basic heuristic exploration can easily cause premature converge and sub-optimal policy. For this reason, we choose to conduct the ablation study on HalfCheetah environment.

In this section, we conduct comprehensive experiments to study how the hyper-parameter $\beta$ could affect the agent's behavior for balancing exploration and exploitation. We studied with following setting: vanilla DDPG baseline \/ L2Explore \/ CMP with only $\mathscr{Q}$ (Meta Actor-Critic), noted as MA2C \/ CMP with $\beta = k $ (noted as CMP-$k$ ) .

\begin{figure*}[t]
    \centering
    \subfigure[Total Return]{
    \label{fig::avg_total_return_a}
    \includegraphics[width=0.32\textwidth]{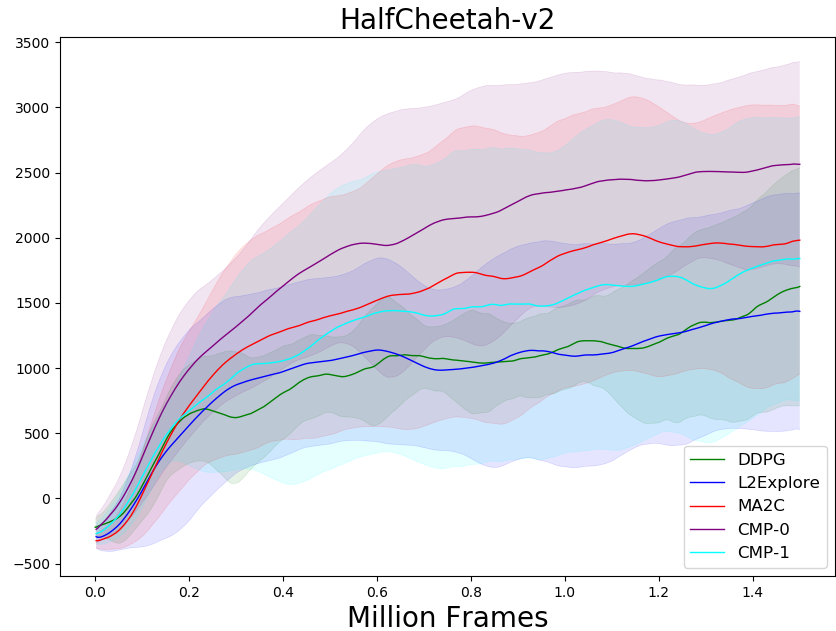}}
    \subfigure[Time Align]{
    \label{fig::avg_total_return_b}
    \includegraphics[width=0.32\textwidth]{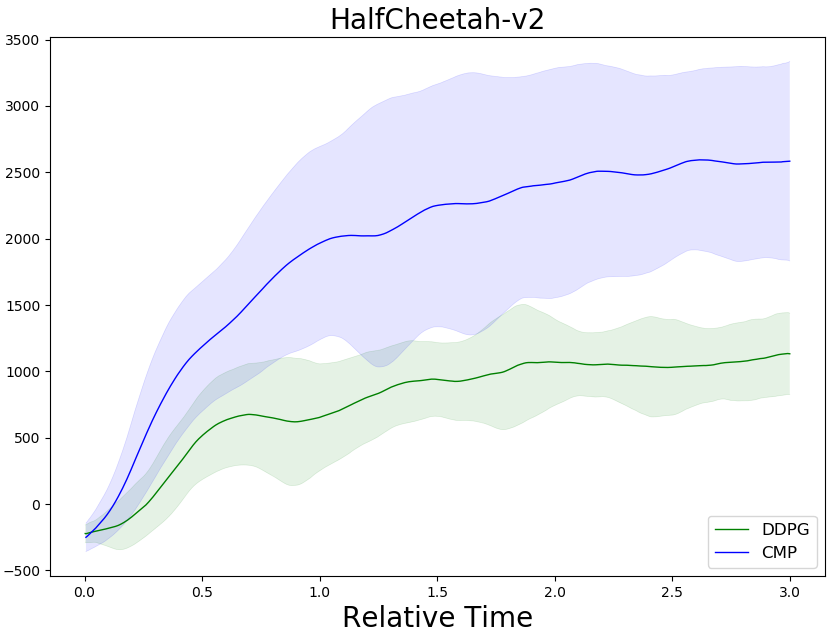}}
    \subfigure[Training Reward]{
    \label{fig::avg_episode_reward}
    \includegraphics[width=0.32\textwidth]{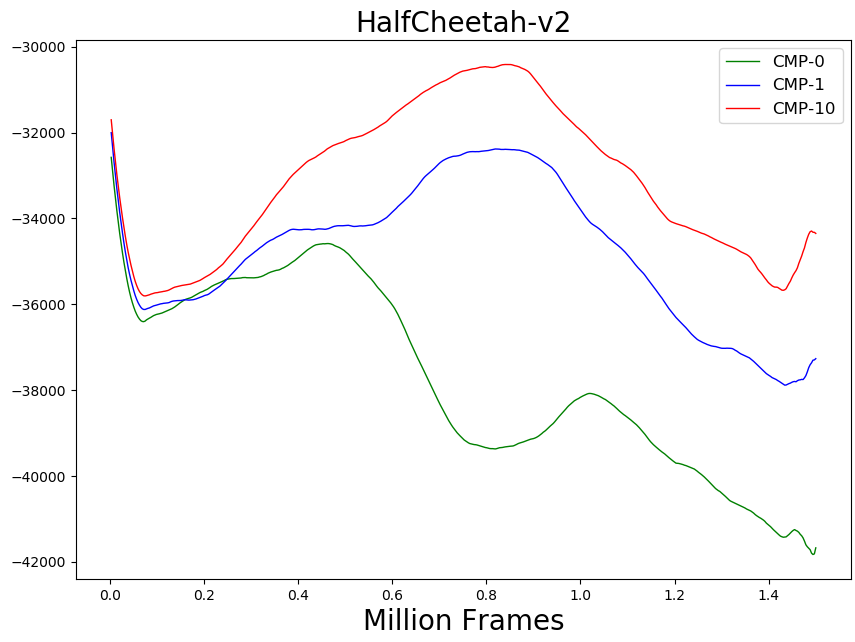}}

    \caption{ Average total return over 100 episodes. (a) Performance Comparison in HalfCheetah: DDPG, L2Explore, MA2C, CMP-0, CMP-1 (b) Wall-block figure: Performance comparison uses the same amount of time and computation resources. (c) Moving sum of training reward collected in exploration over 25 iterations. CMP with different $\beta$  }
\end{figure*}

As shown in Figure. \ref{fig::avg_total_return_a}, compared with MA2C, CMP achieved much better final performance with less samples. CMP methods not only achieve better final performance but also has much better sample efficiency. CMP achieved 1500 with only 0.4M samples which take MA2C about 1.5M samples to achieve. Besides, the comparison between the final performance of  CMP's ( 2562.8 $\pm$ 789 ) and MA2C's( 1981 $\pm$ 1028 ) shows that training with counterfactual mechanism also reduced the variance during training.

The gap between the purple line and the cyan line shows that $\beta$ used to balance the long-term reward and immediate reward do not guarantee final performance improvement. The phenomenon might come from that the q function is more accurate around exploitation policy. In future work, we would analyze the influence of the distance between exploration policy and exploitation policy on agent's final performance.

To further investigate the influence of $\beta$ on agent's behavior, we compared the moving summation of training rewards the agent collected while exploring the environment. As showed in Figure. \ref{fig::avg_episode_reward},  the larger $\beta$ the agent trained with, the more reward is collected during the training phase.

What's more, though CMP spends extra time and resource at each iteration to update the $\pi_e$ and $\hat{\mathscr{Q}}$, CMP finished training in a shorter time. It's because in each iteration, taking Pendulum-v0 as an example, CMP collects 300 samples(100 exploration rollouts and 200 evaluation rollouts) and update the exploitation policy with DDPG for 50 times, while default DDPG updates the policy 50 times every 100 samples. CMP achieved better performance with much less update operation means that CMP obtains much better sample efficiency than the DDPG baseline.

\subsection{ Results on MUJOCO continuous control task }
\begin{figure*}[t]
    \label{fig::MUJOCO}
    \centering
    
    \subfigure{
        \begin{minipage}[t]{0.33\linewidth}
        \centering
        \includegraphics[width=1.1\linewidth]{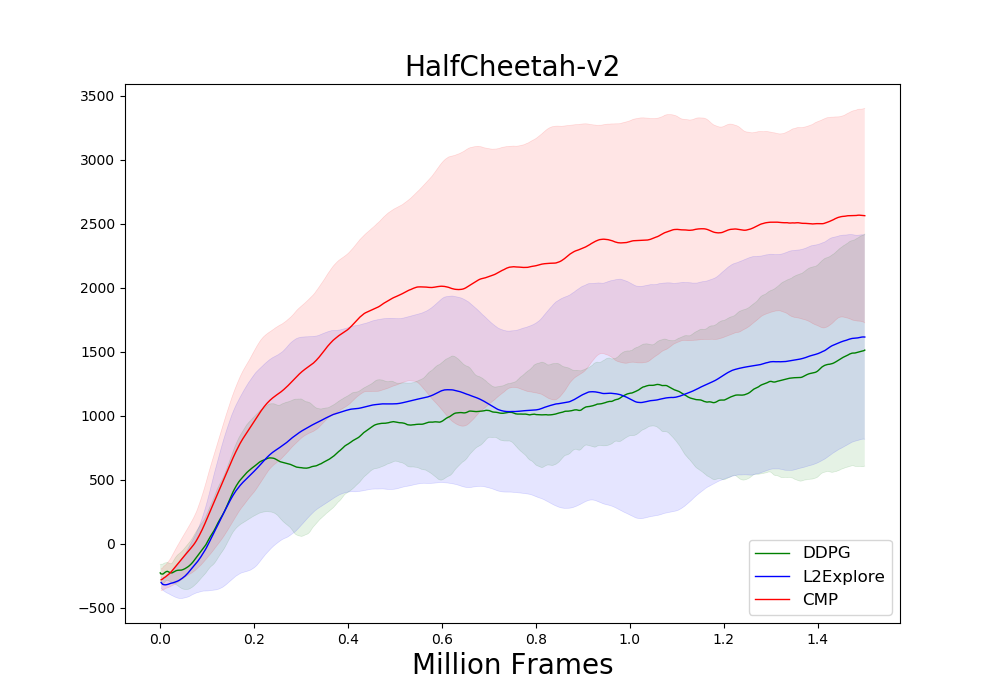}
        \end{minipage}%
    }%
    \subfigure{
        \begin{minipage}[t]{0.33\linewidth}
        \centering
        \includegraphics[width=1.1\linewidth]{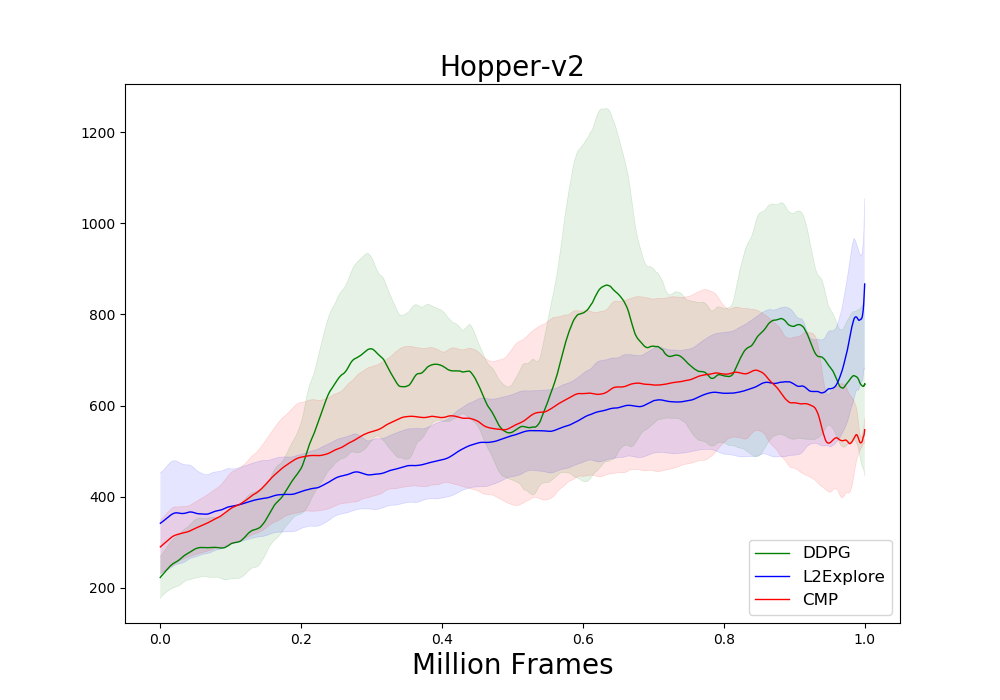}
        \end{minipage}%
    }%
    \subfigure{
        \begin{minipage}[t]{0.33\linewidth}
        \centering
        \includegraphics[width=1.1\linewidth]{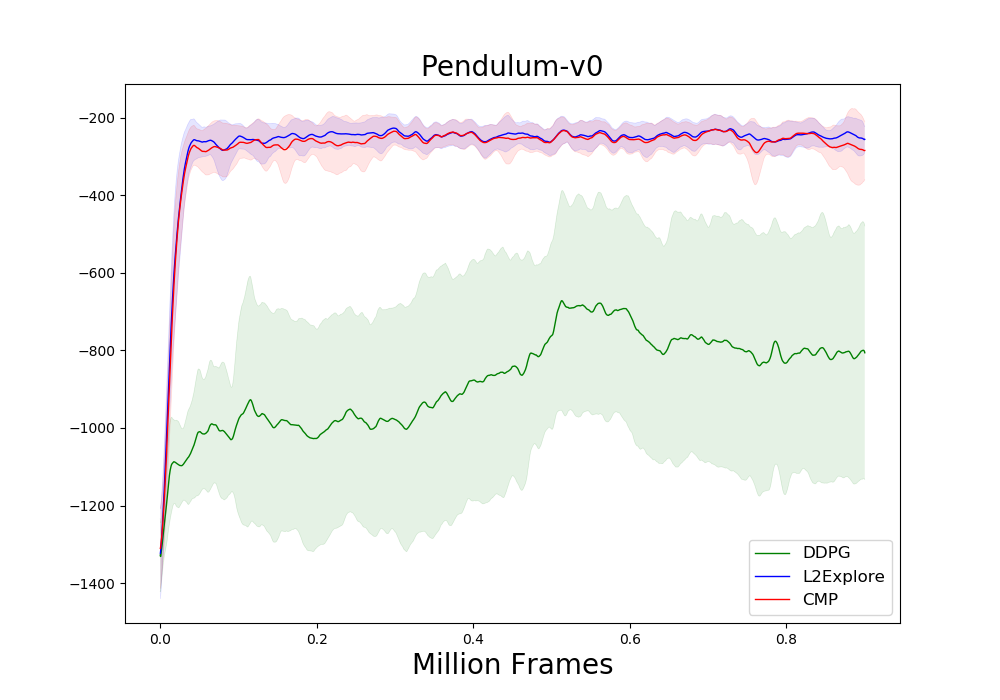}
        \end{minipage}%
    }%
    
    \subfigure{
        \begin{minipage}[t]{0.33\linewidth}
        \centering
        \includegraphics[width=1.1\linewidth]{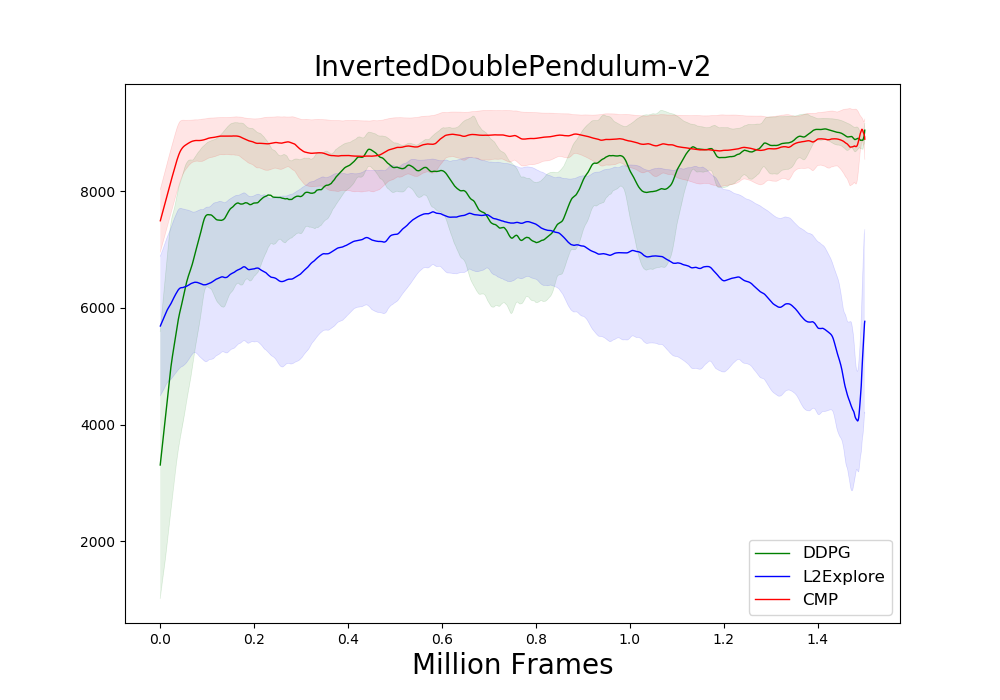}
        \end{minipage}%
    }%
    \subfigure{
        \begin{minipage}[t]{0.33\linewidth}
        \centering
        \includegraphics[width=1.1\linewidth]{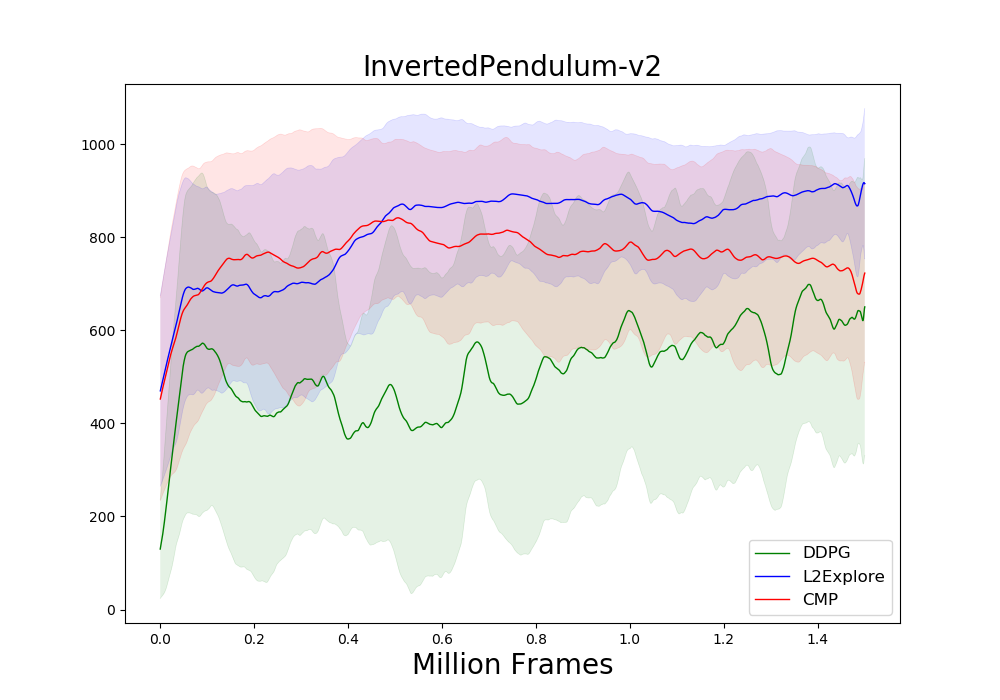}
        \end{minipage}%
    }%
    \subfigure{
        \begin{minipage}[t]{0.33\linewidth}
        \centering
        \includegraphics[width=1.1\linewidth]{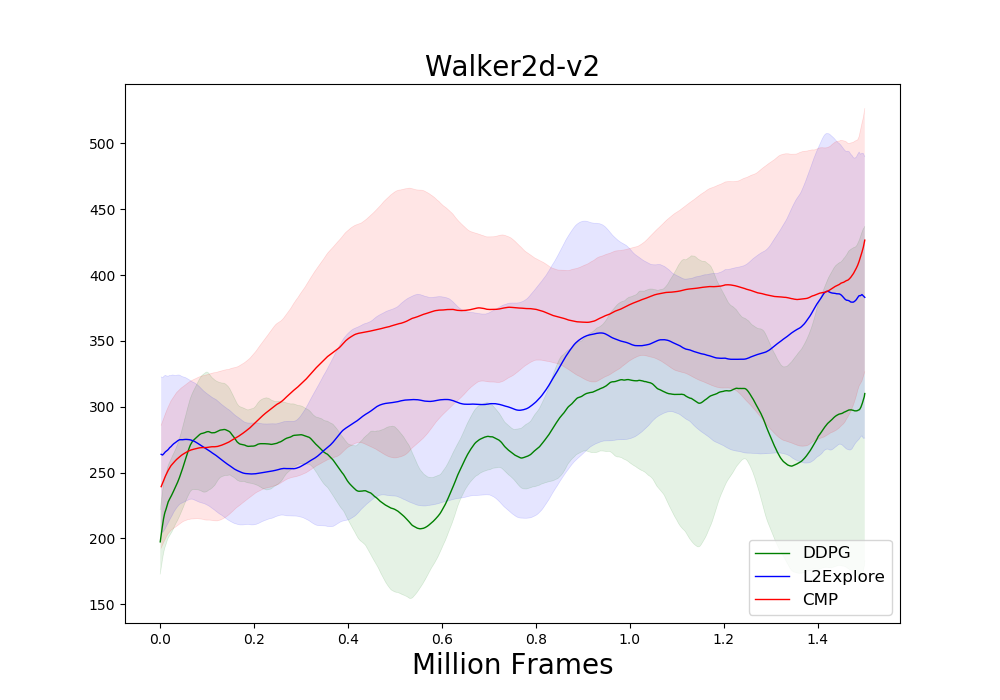}
        \end{minipage}%
    }%
    
    \caption{ Performance Comparison on MUJOCO. Red: CMP(Counterfactual Meta Policy), Blue: L2Explore, Green: DDPG  }
\end{figure*}

The total return of different environments is shown in the Figure. \ref{fig::MUJOCO}. CMP outperforms the DDPG and L2Explore in various continuous control tasks. The solid curves stand for the mean of total return over three seeds and the shaded region corresponds to the standard deviation of total return over 10 seeds.

Though CMP succeeds in many tasks, it's relatively easy to see that CMP is hard to achieve better performance in some environments with these characteristics: 1. the exploration is not the critical problem, 2. The environments have some kind of instability like Hopper-v2, whose length of an episode is unstable. For these environments,  agents need extra samples to learn efficient exploration strategy and easily stuck in the sub-optimal policy. 

\section{Conclusions}
In this paper, we proposed Counterfactual Meta Policy (CMP), a end-to-end meta exploration policy training framework for any off-policy reinforcement methods. With the definition of Counterfactual Value, the learned exploration policy could generated samples that helping the efficiency and efficacy of learning. Empirical results show that our methods effectively learned better exploration strategy that greatly improved learning performance.

\bibliographystyle{unsrt}  
\bibliography{00_bibliography.bib}  

\end{document}